\documentclass{article}

\PassOptionsToPackage{numbers, compress}{natbib}

\usepackage[final]{neurips_2023}

\usepackage[utf8]{inputenc} 
\usepackage[T1]{fontenc}    
\usepackage{hyperref}       
\usepackage{url}            
\usepackage{booktabs}       
\usepackage{amsfonts}       
\usepackage{nicefrac}       
\usepackage{microtype}      
\usepackage{xcolor}         

\usepackage{amsmath}
\usepackage{comment}
\usepackage{graphicx} 
\usepackage{subfigure} 
\usepackage{pdflscape}
\usepackage{algorithm}
\usepackage{algorithmic}
\usepackage{bm}
\usepackage{soul}
\usepackage{listings}
\usepackage{bbm}
\usepackage{multirow}
\usepackage{multicol}

\include{formatting}

\include{xcolor}
\include{xparse}

\NewDocumentCommand{\codeword}{v}{%
\texttt{\textcolor{blue}{#1}}%
}
\newcommand\Tstrut{\rule{0pt}{2.2ex}}         

\title{BayesDLL: Bayesian Deep Learning Library}

\author{
Minyoung Kim$^1$\\
$^1$Samsung AI Center Cambridge, UK\\
{\tt\small mikim21@gmail.com}
\And
Timothy Hospedales$^{1,2}$\\
$^2$University of Edinburgh, UK\\
{\tt\small t.hospedales@ed.ac.uk}
}

\begin{document}

\maketitle

\begin{abstract}
We release a new Bayesian neural network library for PyTorch for large-scale deep networks. Our library implements mainstream approximate Bayesian inference algorithms: variational inference, MC-dropout, stochastic-gradient MCMC, and Laplace approximation. 
The main differences from other existing Bayesian neural network libraries are as follows: 1) Our library can deal with very large-scale deep networks including Vision Transformers (ViTs). 
2) We need virtually zero code modifications for users (e.g., the backbone network definition codes do not neet to be modified at all).
3) Our library also allows the pre-trained model weights to serve as a prior mean, which is very useful for performing Bayesian inference with the large-scale foundation models like ViTs that are hard to optimise from scratch with the downstream data alone. 
Our code is publicly available at: 
\url{https://github.com/SamsungLabs/BayesDLL}\footnote{A mirror repository is also available at: \url{https://github.com/minyoungkim21/BayesDLL}.}. 
\end{abstract}

\section{Bayesian Neural Networks: Overview}\label{sec:bnn}

The followings are the list of approximate Bayesian inference algorithms implemented in the library:
\begin{itemize}
\item Variational Inference (Sec.~\ref{sec:vi})
\item MC-Dropout (Sec.~\ref{sec:mc_dropout})
\item SG-MCMC (SGLD) (Sec.~\ref{sec:sgld})
%
%
\item Laplace Approximation (Sec.~\ref{sec:la})
\end{itemize}

The Bayesian neural network (BNN) is a Bayesian model where we treat the  parameters of the deep neural network (e.g., weights and biases) as {\em random variables} that are endowed with some distribution a priori ({\em prior distribution}). Like other Baysian models, there is a likelihood model that assigns the compatibility score to the observation given the network parameters. 

Formally we use the following notations:
\begin{itemize}
\item $\theta=$ Network parameters (weights $\&$ biases) of the underlying deep model.
\item $\overline{\theta}=$ The most reasonable parameter values before observing any evidence. 
In typical situations, we can have $\overline{\theta}=0$ meaning that we have no prior information, or $\overline{\theta}$ can take {\em pre-trained} model parameters on some base datasets, often called the upstream datasets. For simplicity, we assume a Gaussian prior model in which case $\overline\theta$ becomes the prior mean. More specifically, the prior distribution is written as: 
\begin{align}
p(\theta) = \mathcal{N}(\theta; \overline{\theta}, \sigma^2 I),
\end{align}
where $\sigma^2$ is the prior variance (isotropic Gaussian) chosen by the users. 
Obviously, the prior mean $\overline{\theta}$ and variance $\sigma^2$ are fixed constants.
\item $D=$ Given evidence. Typically $D$ is a supervised dataset ($D=\{(x,y)\}$ where $x \in \mathcal{X}$ is the input and $y \in \mathcal{Y}$ is the target label, either class-valued or real-valued). As conventional practice we have i.i.d.~samples $(x,y) \in D$, which form the likelihood model,
\begin{align}
p(D|\theta) = \prod_{(x,y)\in D} p(y|x,\theta), \ \ \ \ p(y|x,\theta) \propto \exp(-l(x,y;\theta)/\tau),
\end{align}
where $l(x,y;\theta) = l(y, f_\theta(x))$ is the conventional deep learning loss (e.g., cross entropy or $L_2$ distance), $f_\theta(x)\!\in\!\mathcal{Y}$ is the prediction of the deep network with parameters $\theta$ and input $x$, and $\tau$ is the scaling hyperparameter (e.g., temperature for cross entropy in classification cases or variance of the output noise in regression cases). 
\end{itemize}

\paragraph{Neural Network Learning}
The main task is {\em posterior inference}, the task of inferring the posterior distribution $p(\theta|D)$ of the weights given the evidence $D$. That is,
\begin{align}
p(\theta|D) = \frac{p(\theta) \ p(D|\theta)}{\int p(\theta) \ p(D|\theta) \ d\theta}. 
\label{eq:posterior}
\end{align}
The denominator does not in general admit closed-form expression, it is even infeasible to evaluate it exactly. Thus one has to resort to approximation, and several well-known approximate inference algorithms are listed in the beginning of this section, detailed in the next section, and implemented in our \texttt{BayesDLL} library. 

\paragraph{Neural Network Inference} 
Once the posterior inference is done, at test time we can use the posterior to derive the {\em test predictive distribution} $p(y^*|x^*,D)$ where $x^*$ is the test input. In principle,
\begin{align}
p(y^*|x^*,D) = \int p(y^*|x^*,\theta) \ p(\theta|D) \ d\theta.
\label{eq:test_pred}
\end{align}
However, the integration in (\ref{eq:test_pred}) is in general intractable to compute exactly. Instead one can approximate it by the Monte Carlo estimation. If we have a finite number ($S$) of samples from the posterior, then the posterior predictive distribution can be approximated as:
\begin{align}
p(y^*|x^*,D) \approx \frac{1}{S} \sum_{i=1}^S p(y^*|x^*,\theta^{(i)}), \ \ \textrm{where} \ \ \theta^{(i)}\!\sim\!p(\theta|D), \ \ i=1,\dots,S.
\label{eq:predictive_mc}
\end{align}

\begin{figure}
\begin{center}
%
\centering
\includegraphics[trim = 10mm 25mm 5mm 10mm, clip, scale=0.812]{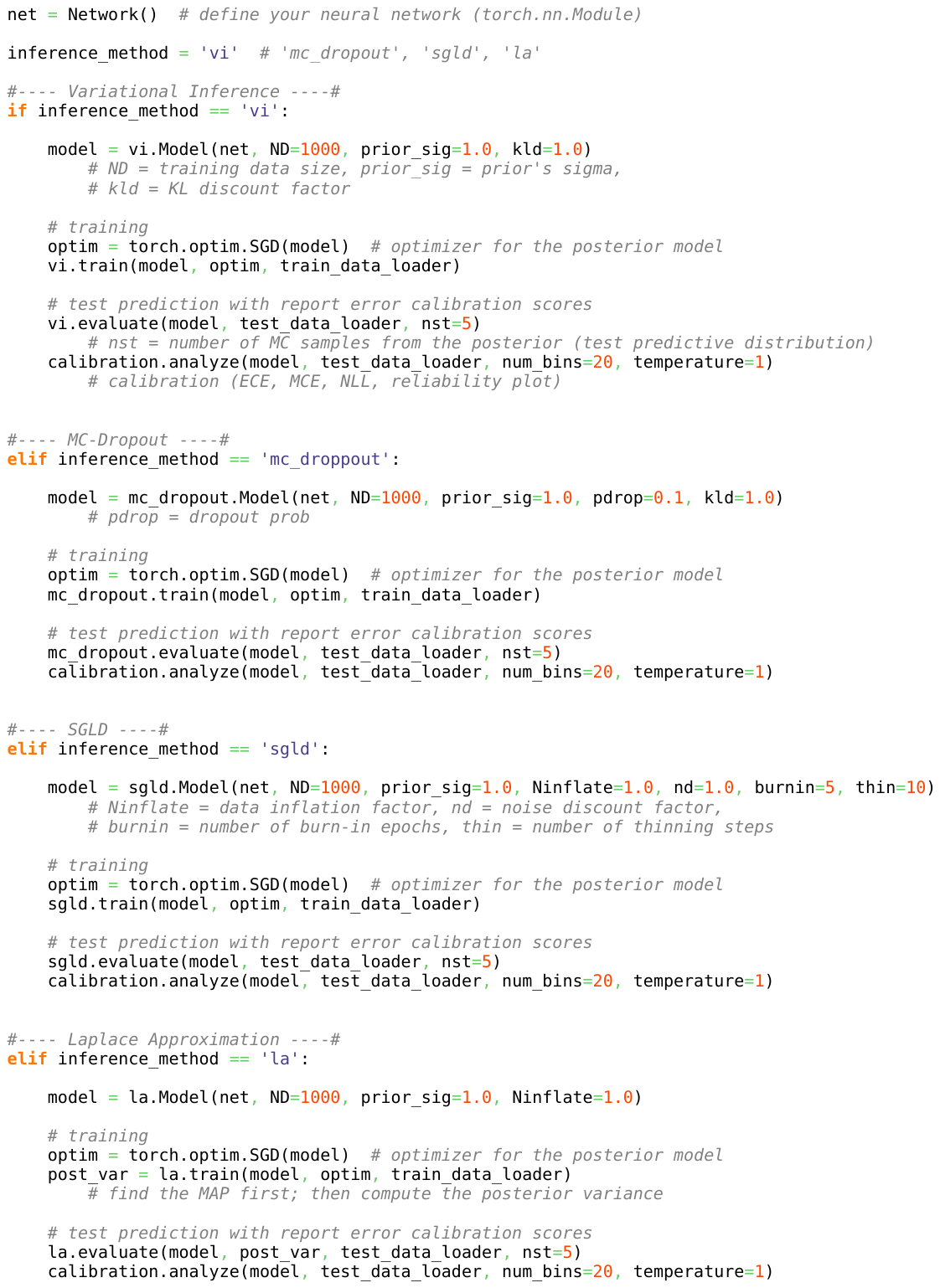}
\end{center}
\caption{BayesDLL usage pseudocodes for the four inference methods. For each method, we have pseudocodes showing how to perform posterior inference (training) and how to obtain a test predictive distribution (test evaluation).
}
\label{fig:pseudocodes}
\end{figure}

\paragraph{BayesDLL Usages (Pseudocodes)} 
The above two steps are implemented in our BayesDLL. For the four inference methods to be described in the next section, we highlight the pseudocodes in Fig.~\ref{fig:pseudocodes}, which shows how to use BayesDLL to do posterior inference and test prediction.

\section{Approximate Inference Algorithms}\label{sec:main}

\subsection{Variational Inference (aka Bayes-by-Backprop~\cite{bayes_by_backprop})}\label{sec:vi}

In the variational inference we typically adopt the following Gaussian\footnote{
Perhaps this assumption/restriction of the tractable density family is one of the main caveats of the variational inference. On the other hand, the MCMC algorithms (e.g., SGLD in Sec.~\ref{sec:sgld}) do not require such an assumption, thus being highly flexible. The only requirement for SGLD is that we can easily (e.g., analytically) compute the gradients of the log-prior $\log p(\theta)$ and the log-likelihood $\log p(y|x,\theta)$ with respect to $\theta$. Sec.~\ref{sec:sgld} for details. 
} densities for both prior and variational posterior:
\begin{itemize}
\item \textbf{Prior:} \ \ $p(\theta) = \mathcal{N}(\theta; \overline\theta, \sigma^2 I)$.
\item \textbf{Variational posterior:} \ \ $q(\theta) = \mathcal{N}(\theta; m, s^2)$ where $s$ is the vectorized standard deviations (of the same shape as $\theta$),  embedded in a diagonal matrix (and $s^2$ means elementwise squaring). Both $m$ and $s$ are the variational parameters to be estimated. Due to the positivity constraint for $s$, we consider positive linking $s = g(\tilde{s})$ where $\tilde{s}$ is free (unconstrained) optimization variables, and $g(\cdot)$ can be typically the exponential function ($s = \exp(\tilde{s})$), the soft-plus function ($s = \log(1+\exp(\tilde{s}))$), or a simple hinge function 
($s = \max(\tilde{s},s_{\min})$ where $s_{\min}$ is a small positive constant such as $10^{-8}$). 
\end{itemize}
The negative ELBO loss function, in the data size normalized and the unbiased minibatch stochastic estimate version, can be written as (here, $B$ denotes a minibatch):
\begin{align}
\textrm{loss} \ = \ \frac{1}{|B|} \sum_{(x,y)\in B} l(x,y; \theta\!=\!m\!+\!\epsilon\!\odot\!s) \ + \ \frac{1}{|D|} \sum_i \frac{1}{2} \bigg( \log \frac{\sigma^2}{s_i^2} - 1 + \frac{s_i^2}{\sigma^2} + \frac{(m_i - \overline{\theta}_i)^2}{\sigma^2} \bigg),
\label{eq:loss_vi}
\end{align}
where $\epsilon \sim \mathcal{N}(0,I)$ and $\odot$ is elementwise product. 
The loss gradient can be easily derived using the chain rule:
\begin{align}
\frac{\partial \textrm{loss}}{\partial m} \ &= \ \frac{1}{|B|} \sum_{(x,y)\in B} \frac{\partial l(x,y; \theta)}{\partial \theta} \ + \ \frac{1}{\sigma^2 |D|} (m-\overline\theta), \\
\frac{\partial \textrm{loss}}{\partial \tilde{s}} \ &= \ \frac{1}{|B|} \sum_{(x,y)\in B} \frac{\partial l(x,y; \theta)}{\partial \theta} \odot \epsilon \odot s \ + \ \frac{1}{|D|} \bigg( \frac{s^2}{\sigma^2}-1 \bigg),
\end{align}
where in this case we assumed the exponential positive linking function ($s = \exp(\tilde{s})$).

Once $m$ and $s$ (that is, $\tilde{s}$) are learned, at test time we can sample $\theta\!\sim\!q(\theta)\!=\!\mathcal{N}(\theta; m, s^2)$. If we consider $S$ samples, then the posterior predictive distribution becomes:
\begin{align}
p(y|x,D) = \frac{1}{S} \sum_{i=1}^S p(y|x,\theta^{(i)}), \ \ \textrm{where} \ \ \theta^{(i)}\!\sim\!\mathcal{N}(\theta; m, s^2), \ \ i=1,\dots,S.
\label{eq:vi_predictive_mc}
\end{align}

\subsection{MC-Dropout}\label{sec:mc_dropout}

In this section we describe {\em our} formulation for the MC-Dropout approximate inference algorithm. This is slightly different from the original version~\cite{mc_dropout} in the following aspects: 1) We allow the Gaussian prior mean is either set to be $0$ (original version) if no prior information is available, or set to be some known values $\overline\theta$ to incorporate the prior knowledge (typically pre-trained network parameters); 2) Whereas the original version dropouts the inputs to layers, we dropout the network parameters instead; Since the former usually requires modification of the network definition codes in order to insert dropout layers, the main advantage of the parameter dropout is that the code modification is not necessary; 3) In the original version the bias parameters take Gaussian posteriors, being different from the mixture of two spiky Gaussian posteriors for the weight parameters; We consider both the Gaussian posterior and the spiky mixture posterior for bias parameters, which is offered as an option for users to select; 4) Moreover, sometimes it is conventional practice not imposing prior for the bias parameters at all, and we incorporate this option as well for the sake of user's convenience. Now we discuss the detailed formulations for our implementation. 

\begin{itemize}
\item \textbf{Prior:} \ \ $p(\theta) = \mathcal{N}(\theta; \overline\theta, \sigma^2 I)$.
\item \textbf{Variational posterior:} \ \ $q(\theta) = \prod_i \big( (1\!-\!p) \cdot \mathcal{N}(\theta_i; m_i, \epsilon^2) + p \cdot \mathcal{N}(\theta_i; \overline\theta_i, \epsilon^2) \big)$, where $\epsilon$ is negligibly small (making two components {\em spiky}), $p \in [0,1]$ corresponds to the dropout probability, and $m$ (of the same shape/size as $\theta$) is the only variational parameters to be estimated. 
\item \textbf{Bias options:} \ \ The bias parameters, denoted by $\theta_b$, can take different prior and/or variational posterior distributions depending on user's option choice. The first option is {\em not imposing prior} on $\theta_b$ at all, more precisely imposing the  uninformative prior $p(\theta_b)\!\propto\!1$, in which case we set $q(\theta_b)$ as a delta function, or virtually equivalent to $q(\theta_b) = \mathcal{N}(\theta_b; m_b, \epsilon^2 I)$, and the consequence is that in the loss function (negative ELBO) we can simply ignore the corresponding KL term; The second option is to place the Gaussian prior $p(\theta_b) = \mathcal{N}(\overline{\theta}_b, \sigma^2 I)$ and Gaussian posterior $q(\theta_b) = \mathcal{N}(\theta_b; m_b, \epsilon^2 I)$, which is exactly the option taken by the original MC-Dropout~\cite{mc_dropout}. This can be implemented by treating the dropout probability $p$ separately for biases (denoted by $p_b$) and non-biases (denoted by $p$), and setting $p_b=0$; And of course the last (default) option is to {\em treat biases in the same way as weights}, in which case we use exactly the above prior and variational posterior.
\end{itemize}

The negative ELBO loss function, in the data size normalized and the unbiased minibatch stochastic estimate version, is comprised of the expected negative log-likelihood (ENLL) $\frac{1}{|B|}\mathbb{E}_q[-\log p(B|\theta)]$ and the KL term $\frac{1}{|D|}\textrm{KL}(q(\theta)||p(\theta))$, where $B$ and $D$ are the minibatch and the whole training set, respectively.
The ENLL term is estimated by Monte Carlo, using the reparametrization trick. We first sample keep-or-dropout binary indicators $z$ (of the same shape/size as $\theta$), specifically $z_i \sim \textrm{Bernoulli}(1\!-\!p)$, where $z_i\!=\!1$ means no-dropout of $\theta_i$ and $z_i\!=\!0$ implies dropout. Then the reparametrized sample is $\theta = z \odot m + (1\!-\!z) \odot \overline\theta$. 
The KL term (between the mixture of Gaussians $q$ and the Gaussian $p$) can be approximated by the same technique as~\cite{mc_dropout}. The final loss function (to be minimized over the variational parameters $m$) is as follows:
\begin{align}
\textrm{loss} \ = \ \frac{1}{|B|} \sum_{(x,y)\in B} l(x,y; \theta = z \odot m + (1\!-\!z) \odot \overline\theta) \ + \ \frac{1}{\sigma^2 |D|} \frac{1\!-\!p}{2} ||m-\overline\theta||_2^2.
\label{eq:loss_mc_dropout}
\end{align}
Note that when there is no prior information on $\theta$, that is, $\overline\theta\!=\!0$, (\ref{eq:loss_mc_dropout}) reduces to the original MC-Dropout. The loss gradient\footnote{We may not need the loss gradient explicitly if one implements it using the parameter-level comp-graph build-up and backprop such as the \texttt{higher} library. We do not utilize this library in our current version (as of August 2023).} can be easily derived using the chain rule:
\begin{align}
\frac{\partial \textrm{loss}}{\partial m} \ = \ \frac{1}{|B|} \sum_{(x,y)\in B} \frac{\partial l(x,y; \theta)}{\partial \theta} \odot z \ + \ \frac{1\!-\!p}{\sigma^2 |D|} (m-\overline\theta),
\end{align}
where $\frac{1}{|B|} \sum_{(x,y)\in B} \frac{\partial l(x,y; \theta)}{\partial \theta}$ can be easily computed by a backprop call provided in most auto-gradient deep learning libraries (e.g., PyTorch or Tensorflow).  

Once $m$ is learned, at test time we can sample $\theta \sim q(\theta)$ from the mixture density. Although this amounts to doing similar dropout sampling used in the ENLL estimation, we often ignore dropout and use the Gaussian sampling. Due to negligible $\epsilon$, we have a {\em deterministic} sample $\theta=m$.

\subsection{SG-MCMC (SGLD)}\label{sec:sgld}

In the stochastic-gradient MCMC (SG-MCMC) approach~\citep{sgld,sgmcmc2,sgmcmc1}, we can collect posterior samples by running a certain stochastic dynamic model whose stationary distribution coincides with the posterior distribution (\ref{eq:posterior}). The stochastic-gradient Langevin dynamic method (SGLD)~\citep{sgld} forms a Langevin dynamic model, 
which amounts to running the following recurrence to collect posterior samples (after some burn-in steps):
\begin{align}
\theta \ \leftarrow \ \theta + \frac{\eta}{2} \nabla \bigg( \log p(\theta) + \frac{|D|}{|B|} \log p(B|\theta) \bigg)
+ \epsilon \sqrt{\eta}
\label{eq:sgld_step}
\end{align}
where $B$ ($\subset\!D$) is a minibatch, $\eta$ is small step size, 
and $\epsilon\!\sim\!\mathcal{N}(0,I)$. 
Note that in the parentheses subject to the derivative, the first log-prior term admits closed-form gradient while the gradient of the second term can be computed by the conventional SGD backprop. Thus each step in (\ref{eq:sgld_step}) is as efficient as the vanilla SGD step.

After a burn-in period, we can maintain those samples $\theta$ to approximate the posterior $p(\theta|D)$. For instance, the running average of the $\theta$ samples, denoted by $\hat{\theta}$, is a good estimate of the mean of the posterior $p(\theta|D)$. 
In the ideal case, we can save all available samples from the posterior (i.e., the iterates from (\ref{eq:sgld_step})), however, due to the large number of parameters in $\theta$, this would easily incur a computational challenge. To this end, in our current implementation\footnote{Alternatively, perhaps more expressive solution might be to estimate a mixture-of-Gaussians density model to fit the posterior samples. We leave this implementation as our future work.} we estimate/maintain the sample means and variances from the posterior samples (via running estimation); and at test time the (approximate) posterior samples are taken from the Gaussian fitted with these sample means and variances.

\subsection{Laplace Approximation}\label{sec:la}

The Laplace approximation essentially approximates the log-posterior $\log p(\theta|D)$ by the second-order Taylor polynomial at the maximum-a-posteriori (MAP) estimate $\theta^*$. More specifically, we first obtain the MAP estimate $\theta^*$ by solving the following optimization problem, typically using the SGD:
\begin{align}
\theta^* = \arg\max_\theta \ \log p(\theta|D) = \arg\max_\theta \ \log p(\theta) + \log p(D|\theta).
\end{align}
Then we approximate $\log p(\theta|D)$ by the quadratic Taylor polynomial at $\theta=\theta^*$, which is simplified as follows due to the vanishing gradient at the (local) optimum (i.e., $\nabla\log p(\theta^*|D) = 0$):
\begin{align}
\log p(\theta|D) \ \approx \ \frac{1}{2} (\theta-\theta^*) \nabla^2 \log p(\theta^*|D) (\theta-\theta^*) + \textrm{const.}
\label{eq:la_log_posterior}
\end{align}
Assuming that the Hessian $\nabla^2 \log p(\theta^*|D)$ is negative definite\footnote{If not, one can perform the Generalised Gauss-Newton (GGN) approximation for the Hessian~\cite{Schraudolph02,Graves11,Martens14}. However, we omit this step for simplicity, and 
our diagonal empirical Fisher approximation in (\ref{eq:la_posterior_emp_fisher}) implicitly handles this potential issue. 
}, the equation (\ref{eq:la_log_posterior}) essentially leads to the Gaussian posterior approximation,
\begin{align}
p(\theta|D) \ \approx \ \mathcal{N}\big( \theta; \theta^*, -( \nabla^2 \log p(\theta^*|D) )^{-1} \big).
\label{eq:la_posterior}
\end{align}

Here arises the infamous computational challenge in Hessian evaluation and inversion from (\ref{eq:la_posterior}). First, to circumvent the $O(d^2)$ memory overhead for saving the Hessian matrix and prohibitive $O(d^3)$ matrix inversion time, where $d=\dim(\theta)$, we consider the diagonal Hessian approximation. 
Secondly, to deal with the overhead of the Hessian computation, we adopt the famous {\em empirical Fisher information approximation} for the Hessian.  
For concreteness, we here derive the details of the diagonal empirical Fisher approximation. Letting the training data $D=\{(x_n,y_n)\}_{n=1}^N$, 
\begin{align}
\nabla^2 \log p(\theta^*|D) = \nabla^2 \log p(\theta^*) + \sum_{n=1}^N \nabla^2 \log p(y_n|x_n,\theta^*)
\label{eq:la_deriv_1}
\end{align}
The second term in the RHS of (\ref{eq:la_deriv_1}) can be approximated by the empirical Fisher information as (\ref{eq:la_deriv_2}), which is essentially obtained by replacing the model distribution $p(y|x,\theta)$ in the Fisher information by the plug-in estimate or the empirical distribution $\frac{1}{N} \sum_{n=1}^N \delta(y\!-\!y_n|x_n)$:
\begin{align}
\nabla^2 \log p(\theta^*|D) \approx \nabla^2 \log p(\theta^*) - \sum_{n=1}^N \nabla \log p(y_n|x_n,\theta^*) \nabla \log p(y_n|x_n,\theta^*)^\top.
\label{eq:la_deriv_2}
\end{align}
Now, we further approximate the dyads by the diagonal matrix (i.e., element-wise squaring instead of outer product), leading to:
\begin{align}
\nabla^2 \log p(\theta^*|D) \approx \nabla^2 \log p(\theta^*) - \sum_{n=1}^N \textrm{Diag}\Big( \nabla \log p(y_n|x_n,\theta^*)^2 \Big),
\label{eq:la_deriv_3}
\end{align}
where the squaring in (\ref{eq:la_deriv_3}) is element-wise, and $\textrm{Diag}(v)$ is the diagonal matrix with the vector $v$ embedded in the diagonal entries. 
Lastly, assuming isotropic Gaussian prior $p(\theta) = \mathcal{N}(\theta; \overline{\theta}, \sigma^2 I)$, we have the final posterior approximation:
\begin{align}
p(\theta|D) \ \approx \ \prod_i \mathcal{N}( \theta_i; \theta^*_i, v_i) \ \ \textrm{where} \ \ v_i = \frac{1}{
\frac{1}{\sigma^2} + \sum_{n=1}^N \big[\nabla \log p(y_n|x_n,\theta^*)\big]_i^2
}.
\label{eq:la_posterior_emp_fisher}
\end{align}
Now (\ref{eq:la_posterior_emp_fisher}) can be computed with one forward-pass for each data instance, and all operations are done in $O(d)$ time/memory.

Although there exist other Hessian approximation strategies, notably the block diagonal approximation schemes such as the Kronecker factorization~\cite{kfac_la}, they can potentially introduce considerable computational overhead compared to the diagonal one, which often hinders their applications to the large-scale networks such as Vision Transformers. For this reason we omit the implementation of these methods in our library.




\section{Uncertainty Quantification}\label{sec:uncertainty}

One of the key benefits of using Bayesian deep models is its capability of capturing uncertainty in their predictive distributions. There are two popular types of methods to quantify/measure how well the uncertainty is captured in the underlying models: error calibration and negative log-likelihood.

\subsection{Error Calibration}\label{sec:calibration}

The popular error calibration metrics such as ECE and MCE~\cite{ece} as well as the visualization tool like the Reliability plot~\cite{reliability1, reliability2} belong to this category.

The key idea is to measure {\em how well the prediction accuracy and the prediction confidence are aligned}. Most approaches rely on metric evaluation based on confidence binning. More specifically, assume that we have class predictions by the model $p(y=j|x^i)$ for $i=1,\dots,N$ and $j=1,\dots,K$ where $K$ is the class cardinality. Let $y^\textrm{true}(x)$ be the ground-truth class label. We consider bin size $M$ (bin index $m=0,1,\dots,M\!-\!1$). 
\begin{itemize}
\item Initialize: $binsize[0:M] = acc[0:M] = conf[0:M] = 0$
\item For $i=1,\dots,N$ and $j=1,\dots,K$:
\item[] \ \ \ \ \ \ - Determine $m = $ Bin ID that $p(y=j|x^i)$ belongs to
\item[] \ \ \ \ \ \ - $binsize[m] \leftarrow binsize[m] + 1$
\item[] \ \ \ \ \ \ - $acc[m] \leftarrow acc[m] +  I(y^\textrm{true}(x_i)=j)$
\item[] \ \ \ \ \ \ - $conf[m] \leftarrow conf[m] + p(y=j|x^i)$
\end{itemize}

It would be desirable to have a prediction model that leads to $acc[m] \approx conf[m]$ for all confidence level $m=0,1,\dots,M-1$. There are several ways to visualize or quantify this goodness of alignment.

\textbf{Reliability plot} is just a simple plot of $acc[0:M]$ (Y-axis) vs.~$conf[0:M]$ or bin centers (X-axis). Thus in the ideal case (0 calibration error), this plot would coincide with $Y=X$ line.

\textbf{ECE and MCE} can be computed by the following formulas:
\begin{align}
\textrm{ECE} &= \ \frac{1}{M} \sum_{m=0}^{M-1} \frac{binsize[m]}{N\cdot K} \cdot \Big| acc[m] - conf[m] \Big| \\
\textrm{MCE} &= \ \arg\max_{0\leq m < M} \Big| acc[m] - conf[m] \Big|
\end{align}

\textbf{Temperature scaling:} We typically have a logit vector $f(x)\in\mathbb{R}^K$ as an output of the neural network, before soft-max normalizing it to $p(y|x)$. We consider the (temperature) scaling of this logit before soft-max, that is,
\begin{align}
p_T(y|x) = \frac{e^{f_y(x)/T}}{\sum_{j=1}^K e^{f_j(x)/T}}.
\end{align}
Obviously $T\!=\!1$ is the default setting, but one can find the best $T$ that minimizes the calibration error. To this end, by regarding $T$ as an optimization parameter, we typically form a maximum likelihood estimation problem on the validation set. More specifically,
\begin{align}
\min_T \ \mathbb{E}_{(x,y)\sim D_V} [-\log p_T(y|x)],
\end{align}
where $D_V$ is the validation data set.
Once the optimal $T$ is found, we can report the {\em temperature-scaled} calibration error metrics with $p_T(y|x)$.

\begin{figure}
\begin{center}
%
\centering
\includegraphics[trim = 0mm 0mm 0mm 0mm, clip, scale=0.248]{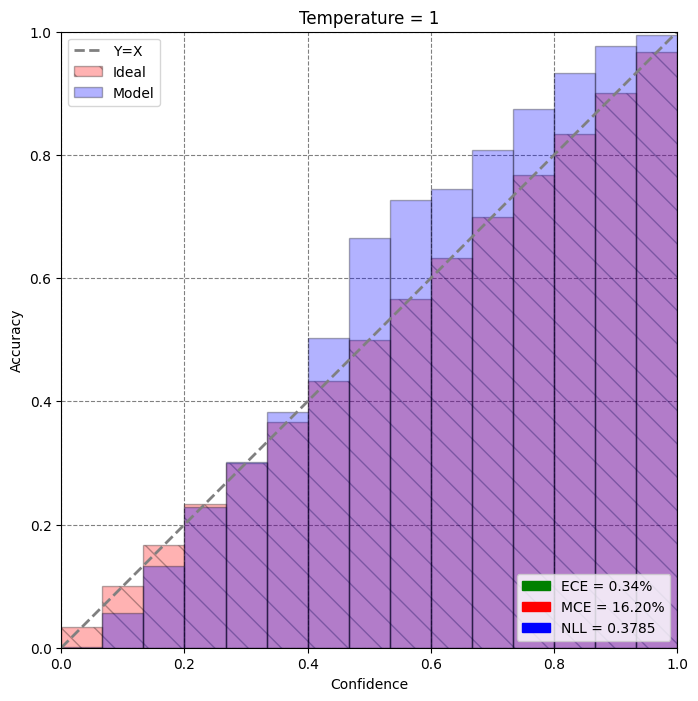} \ 
\includegraphics[trim = 0mm 0mm 0mm 0mm, clip, scale=0.248]{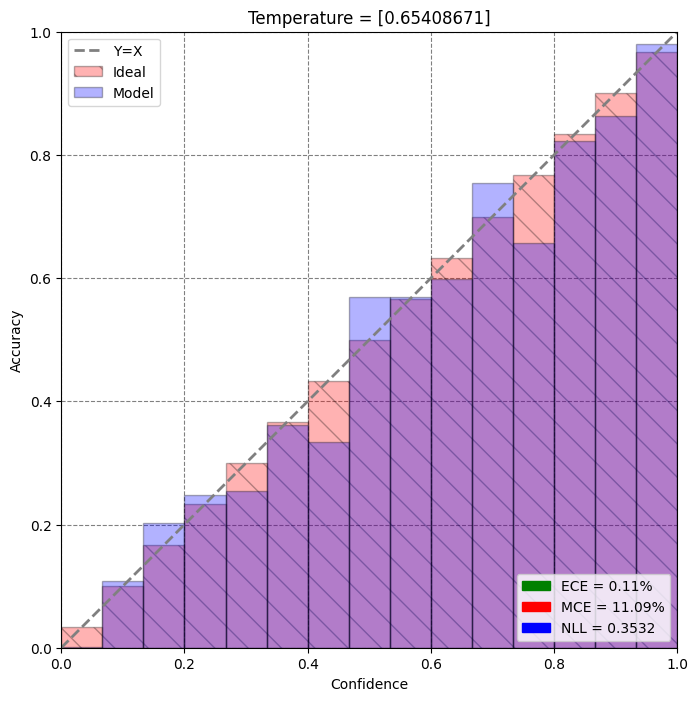} \ 
\includegraphics[trim = 0mm 0mm 0mm 0mm, clip, scale=0.295]{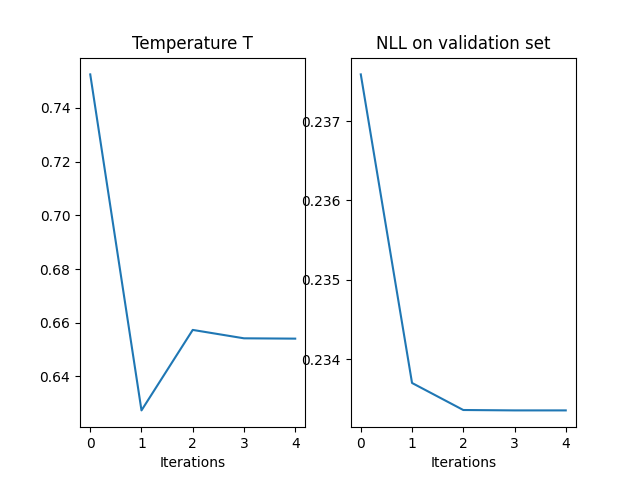}
\\
(a) Reliability plot ($T=1$) \ \ \ \ \ \ \ \ \ \ (b) Reliability plot ($T^*$) \ \ \ \ \ \ \ \ \ \ (c) Temp.~scaling optim. curve
\end{center}
\vspace{-0.5em}
\caption{Our \texttt{BayesDLL} library can produce reliability plots during model training. (a) Reliability plot with default temperature $T\!=\!1$, (b) Reliability plot after temperature scaling, and (c) Temperature scaling optimisation learning curve. In the reliability plots we also show ECE, MCE, and NLL metrics on the lower-right corner. 
}
\label{fig:uq}
\end{figure}

Our \texttt{BayesDLL} library can produce reliability plots and ECE/MCE metrics during model training. See Fig.~\ref{fig:uq} for the examples.

\subsection{Negative Log-Likelihood}\label{sec:nll}

The negative log-likelihood (NLL) on the test data set is the standard statistical metric that measures how close the model's predictive distribution is to the true labeling distribution. It can be computed as follows:
\begin{align}
\textrm{NLL} = \mathbb{E}_{(x,y)\sim D_T} [-\log p(y|x)],
\end{align}
where $D_T$ is the test data set.

\section{Implementation Notes}\label{sec:impl_notes}

Our current library implements four different Bayesian deep learning methods as well as the baseline deterministic (non-Bayesian) method. Which method is used can be specified by the flag \texttt{-{}-method}. For instance, one can add \texttt{-{}-method mc\_dropout} flag in the command line. Within each method, we also list the {\em method-specific} hyperparameters. 
\begin{itemize}
\item \texttt{"vanilla"}: This is a vanilla deterministic deep learning, aka SGD (stochastic gradient descent) learning. We also allow {\em weight decay} to L2-penalise deviation from the pre-trained parameters or zero parameters, as well as the bias option for the L2 penalty.
  \begin{itemize}
      \item \texttt{wd (eg, 1e-4)}: The weight decay (L2 regularisation) coefficient. The L2 penalty is measured based on either deviation from the pre-trained parameters or from 0. 
      \item \texttt{bias $\{$"penalty","ignore"$\}$}: How to treat the bias parameters in L2 penalty. \texttt{"penalty"} specifies the same treatment as weight parameters, while \texttt{"ignore"} simply ignores the bias deviation (analogous to uninformative bias prior in Bayesian methods).
  \end{itemize}
\item \texttt{"vi"}: This is the variational inference method. The related hyperparameters are as follows:
  \begin{itemize}
      \item \texttt{prior\_sig (eg, 0.01)}: This specifies the standard deviation $\sigma$ of the prior Gaussian distribution. 
      \item \texttt{bias $\{$"informative","uninformative"$\}$}: How to treat the bias parameters in the prior. \texttt{"informative"} specifies the same treatment as weight parameters, while \texttt{"uninformative"} simply adopts $p(\theta_b) \propto 1$. This amounts to dropping the KL terms for the bias parameters. 
      \item \texttt{kld (eg, 0.1)}: The discount factor for the KL term. The KL term is multiplied by this factor. This is related to the training data size inflation due to data augmentation.
      \item \texttt{nst (eg, 5)}: The number of posterior samples at test time (i.e., $S$ in (\ref{eq:vi_predictive_mc})).
  \end{itemize}
\item \texttt{"mc\_dropout"}: This is the MC-Dropout, and the related hyperparameters are as follows:
  \begin{itemize}
      \item \texttt{prior\_sig (eg, 0.01)}: This specifies the standard deviation $\sigma$ of the prior Gaussian distribution. 
      \item \texttt{p\_drop (eg, 0.1)}: This specifies the dropout probability. 
      \item \texttt{bias $\{$"gaussian","spikymix", "ignore"$\}$}: How to treat the bias parameters in the prior. \texttt{"gaussian"} takes Gaussian $q(\theta_b)$, thus no dropout for bias parameters; \texttt{"spikymix"} specifies the same treatment as weight parameters; while \texttt{"ignore"} simply ignores prior and posterior for bias parameters. 
      \item \texttt{kld (eg, 0.1)}: The discount factor for the KL term. The KL term is multiplied by this factor. This is related to the training data size inflation due to data augmentation.
      \item \texttt{nst (eg, 5)}: This is the number of posterior samples to be sampled at test time (i.e., $S$ in (\ref{eq:vi_predictive_mc})).
  \end{itemize}
\item \texttt{"sgld"}: This is the SGLD whose related hyperparameters are as follows:
  \begin{itemize}
      \item \texttt{prior\_sig (eg, 0.01)}: This specifies the standard deviation $\sigma$ of the prior Gaussian distribution. 
      \item \texttt{Ninflate (eg, 1e3)}: Data inflation factor (due to data augmentation). The training data size $|D|$ is inflated by this factor. 
      \item \texttt{nd (eg, 0.1)}: Noise discount factor. The noise term in the SGLD iteration is multiplied by this factor. 
      \item \texttt{burnin (eg, 20)}: Burn-in period (in epochs).
      \item \texttt{thin (eg, 10)}: Thinning steps (in batch iterations). 
      \item \texttt{bias $\{$"informative","uninformative"$\}$}: How to treat the bias parameters in the prior. \texttt{"informative"} specifies the same treatment as weight parameters, while \texttt{"uninformative"} simply adopts $p(\theta_b) \propto 1$. This amounts to dropping the prior term in the SGLD iteration. 
      \item \texttt{nst (eg, 5)}: The number of posterior samples to be sampled at test time (i.e., $S$ in (\ref{eq:vi_predictive_mc})). Recall that in the current version we use a sample-estimated Gaussian for the posterior approximation. Thus this is the number of Gaussian samples. 
  \end{itemize}
\item \texttt{"la"}: This is the Laplace approximation, and the related hyperparameters are as follows:
  \begin{itemize}
      \item \texttt{prior\_sig (eg, 0.01)}: This specifies the standard deviation $\sigma$ of the prior Gaussian distribution. 
      \item \texttt{Ninflate (eg, 1e3)}: Data inflation factor (due to data augmentation). The training data size $|D|$ is inflated by this factor. 
      \item \texttt{bias $\{$"informative","uninformative"$\}$}: How to treat the bias parameters in the prior. \texttt{"informative"} specifies the same treatment as weight parameters, while \texttt{"uninformative"} simply adopts $p(\theta_b) \propto 1$. This amounts to dropping the prior terms for bias parameters in the MAP objective. 
      \item \texttt{nst (eg, 5)}: The number of posterior samples at test time (i.e., $S$ in (\ref{eq:vi_predictive_mc})).
  \end{itemize}
\end{itemize}


\section{Experiments}\label{sec:expmt}

In this section we demonstrate the execution results of training and testing with our Bayesian neural network library. In Sec.~\ref{sec:mnist} we provide extensive comparison among various running options and hyperparameters, especially the number of posterior samples during test prediction (\codeword{nst}), how to treat the bias parameters (\codeword{bias}, e.g., either not imposing a prior or treating the same way as weight parameters), the choice of prior scale (\codeword{prior_sig}), and so on. 
In Sec.~\ref{sec:large_scale} we test our library on large-scale neural networks for vision tasks, in particular, ResNet-101 and Vision Transformer (ViT) models. For these models, we will show that learning the models from  scratch (i.e., uninformative zero-mean Gaussian prior $p(\theta)$) often fails. Instead, we impose a prior that is centered at the pre-trained model parameters that are available publicly, which leads to prediction performance comparable to conventional warm-start/fine-tuning deterministic model learning, but with better uncertainty calibration.

\subsection{Testing Various Options/Hyperparameters with MLP on MNIST}\label{sec:mnist}

\textbf{Experimental setup.} 
The original MNIST training dataset is randomly split into $50\%$ training and $50\%$ validation sets, where the latter is used to determinne early stopping of training iterations. The neural network we adopted is a fully-connect network (aka, MLP), which has three hidden layers with 1000 units, followed by the final linear prediction head. For the nonlinearity the ReLU activation is used. For all competing methods, we set the maximum training epochs as 100, learning rate $10^{-2}$, batch size 128, SGD optimizer with momentum 0.5. All training starts with randomly initialized model parameters.

\textbf{Competing approaches.} 
\texttt{Vanilla} is the non-Bayesian deterministic SGD learning, where optionally we can impose the L2 regularization via the weight-decay option (either wd=0 or wd=$10^{-4}$). 
\texttt{VI} stands for variational inference. We take as default hyperparameters the prior scale $\sigma=1.0$ (except $\sigma=0.01$ for \texttt{Laplace}) and the KL discount factor $10^{-3}$.
For \texttt{MC-Dropout}, we use the default dropout probability $0.1$ while the prior scale and the KL discount factor have the same default values as \texttt{VI}. Recalling from Sec.~\ref{sec:mc_dropout}, there are three bias treatment options, which are abbreviated as: \texttt{ga} (Gaussian prior, also conforming to the original version), \texttt{sm} (the spiky mixture prior, thus the same treatment as weight parameters), and uninformative prior. 
In \texttt{SGLD}, we take the burn-in steps for the first 5 epochs, followed by thinning at every 10 batch iterations. The training data inflation factor is set to $10^3$. 
Lastly, for \texttt{Laplace} approximation, we use the same inflation factor, but the prior scale is set to $\sigma=0.01$ since having larger scale (e.g., $\sigma=1.0$) led to failure in all cases. 


\begin{table}[t!]
\setlength{\tabcolsep}{2.5pt}
\caption{MNIST test prediction errors ($\%$). For the Bayesian neural network models, we mainly vary the bias treatment option, either imposing informative bias prior (the same treatment as weight parameters) or uninformative (essentially, not imposing prior on the bias parameters). The other hyperparameters are roughly default values (See text for details), and the test-time posterior sampling is not done, but the posterior mean is used. 
}
\vspace{-0.2em}
\centering
\begin{footnotesize}
\centering
\begin{tabular}{ccccccccccccc}
\toprule
\multirow{2}{*}{} & 
\multicolumn{3}{c}{Vanilla} &
\multicolumn{2}{c}{VI} &
\multicolumn{3}{c}{MC-Dropout} &
\multicolumn{2}{c}{SGLD} &
\multicolumn{2}{c}{Laplace} 
\\
\cmidrule(lr){2-4} \cmidrule(lr){5-6} \cmidrule(lr){7-9} \cmidrule(lr){10-11} \cmidrule(lr){12-13}
 & wd=0 & bias=1 & bias=0 & bias=1 & bias=0 & bias=ga & bisa=sm & bias=0 & bias=1 & bias=0 & bias=1 & bias=0 
\\
\hline
Error ($\%$)\Tstrut & 2.52 & 2.55 & 2.50 & 2.25 & 2.43 & 2.47 & 2.47 & 2.48 & 2.74 & 2.73 & 2.41 & 2.53 \\
\bottomrule
\end{tabular}
\end{footnotesize}
\label{tab:mnist_main}
\end{table}

\textbf{Results on prediction errors.}
As shown in Table~\ref{tab:mnist_main}, all methods perform equally well whereas \texttt{SGLD} slightly falls short. Overall the different bias treatment options have little impact on the prediction performance.

\textbf{Impact of the number of posterior samples at test prediction.}
In the Bayesian neural networks, one can incorporate the prediction uncertainty by marginalizing over the posterior samples, which is in practice typically done by Monte-Carlo averaging such as (\ref{eq:vi_predictive_mc}) for \texttt{VI} and similarly for other methods. This is known to improve the uncertainty calibration, i.e., consistency between prediction confidence and accuracy. We check this property by comparing two different settings, the number of posterior samples \codeword{nst}$=0$ (using the posterior mean) and \codeword{nst}$=5$.
As shown in Table~\ref{tab:mnist_nst}, increasing the number of posterior samples in test prediction leads to reduction in ECE and MCE metrics, indicating that the models are better calibrated.

\begin{table}[t!]
\setlength{\tabcolsep}{5.7pt}
\caption{Impact of the number of posterior samples at test prediction in MNIST. 
The number of posterior samples nst=0 amounts to using the posterior mean.
}
\vspace{-0.2em}
\centering
\begin{footnotesize}
\centering
\begin{tabular}{ccccccccccc}
\toprule
\multirow{2}{*}{} & 
\multicolumn{2}{c}{Vanilla} &
\multicolumn{2}{c}{VI} &
\multicolumn{2}{c}{MC-Dropout} &
\multicolumn{2}{c}{SGLD} &
\multicolumn{2}{c}{Laplace} 
\\
\cmidrule(lr){2-3} \cmidrule(lr){4-5} \cmidrule(lr){6-7} \cmidrule(lr){8-9} \cmidrule(lr){10-11}
 & wd=0 & wd=$10^{-4}$ & nst=0 & nst=5 & nst=0 & nst=5 & nst=0 & nst=5 & nst=0 & nst=5 
\\
\hline
Error ($\%$)\Tstrut & 2.52 & 2.55 & 2.25 & 2.75 & 2.47 & 2.43 & 2.74 & 2.75 & 2.41 & 1.51
\\
ECE ($\%$)\Tstrut & 0.22 & 0.20 & 0.12 & 0.11 & 0.32 & 0.18 & 0.23 & 0.21 & 0.24 & 0.12
\\
MCE ($\%$)\Tstrut & 22.04 & 17.33 & 14.64 & 11.71 & 21.83 & 11.19 & 13.76 & 13.56 & 17.82 & 8.04
\\
NLL ($\times 10^{-2}$)\Tstrut & 9.54 & 9.38 & 8.11 & 8.57 & 11.28 & 8.75 & 10.00 & 9.27 & 9.45 & 9.36
\\
\bottomrule
\end{tabular}
\end{footnotesize}
\label{tab:mnist_nst}
\end{table}

\textbf{Impact of prior scale ($\sigma$).}
We test how the Bayesian models behave when we change the prior scale. From the default value $\sigma=1.0$, we reduce it to $\sigma=0.01$. As the results in Table~\ref{tab:mnist_prior_sig}, the prediction errors barely change, but there are slight improvement in the uncertainty calibration scores. This may be attributed to the stronger regularisation effect, where further deviation from the prior mean $0$ weight parameters is penalised more severely. 

\begin{table}[t!]
\setlength{\tabcolsep}{1.6pt}
\caption{Impact of the prior scale ($\sigma$) in MNIST. In the table entries, ``$a\!\to\!b$'' indicates that score $a$ with $\sigma=1.0$ is changed to score $b$ with $\sigma=0.01$. 
}
\vspace{-0.2em}
\centering
\begin{footnotesize}
\centering
\begin{tabular}{ccccccc}
\toprule
\multirow{2}{*}{} & 
\multicolumn{2}{c}{VI} &
\multicolumn{2}{c}{MC-Dropout} &
\multicolumn{2}{c}{SGLD}
\\
\cmidrule(lr){2-3} \cmidrule(lr){4-5} \cmidrule(lr){6-7} 
 & nst=0 & nst=5 & nst=0 & nst=5 & nst=0 & nst=5 
\\
\hline
Error ($\%$)\Tstrut & $2.25\!\to\!2.26$ & $2.75\!\to\!2.63$ & $2.47\!\to\!2.42$ & $2.43\!\to\!2.52$ & $2.74\!\to\!2.70$ & $2.75\!\to\!2.70$ 
\\
ECE ($\%$)\Tstrut & $0.12\!\to\!0.12$ & $0.11 \!\to\!0.10$ & $0.32\!\to\!0.32$ & $0.18\!\to\!0.17$ & $0.23\!\to\!0.18$ & $0.21\!\to\!0.16$ 
\\
MCE ($\%$)\Tstrut & $14.64\!\to\!11.50$ & $11.71\!\to\!10.50$ & $21.83\!\to\!21.72$ & $11.19\!\to\!22.27$ & $13.76\!\to\!11.95$ & $13.56\!\to\!12.56$ 
\\
NLL ($\times 10^{-2}$)\Tstrut & $8.11\!\to\!7.75$ & $8.57\!\to\!8.44$ & $11.28\!\to\!11.24$ & $8.75\!\to\!8.50$ & $10.00\!\to\!9.49$ & $9.27\!\to\!9.42$ 
\\
\bottomrule
\end{tabular}
\end{footnotesize}
\label{tab:mnist_prior_sig}
\end{table}

\subsection{Large-Scale Backbones including Foundation Models}\label{sec:large_scale}

Next we test our library on the large-scale backbone networks. We consider two popular deep networks for vision tasks: \textbf{ResNet-101} and \textbf{Vision Transformer (ViT)} specifically the version known as \textbf{ViT-L-32}, where the former consists of about $43$ million parameters and the latter about $305$ million parameters. 
For simplicity we consider the image classification vision tasks with the Pets~\cite{pets} and Flowers~\cite{flowers} datasets that contain images of 37 and 102 different categories, respectively. For Pets, we randomly split the official training data into $50\%$ training and $50\%$ validation sets.
For Flowers, we merge the official training and validation data splits, and randomly split them into $50\%$ training and $50\%$ validation sets.

As it is widely believed that training such large-scale networks from the scratch is very difficult and often leads to inferior solutions, we instead adopt the pre-trained model weights in the form of prior mean parameters in the Bayesian models. That is, instead of having $0$-mean prior (i.e., $\overline{\theta}=0$) as usual practice, we set the prior mean equal to the pre-trained weights\footnote{We simply employ the network architecture definitions and the network weights obtained from pre-training with the ImageNet subsets~\cite{deng2009imagenet}, available at \url{https://pytorch.org/vision/main/models.html}. Note that this feature of flexible external code incorporation, without any modification of the original code, is one of the key benefits of the proposed library.}. In our quick experiments with the (non-Bayesian) SGD learning (denoted by \texttt{Vanilla}) in Table~\ref{tab:nopretr_vs_pretr}, we can verify that there is huge performance difference between the trained models with and without pre-trained weights, signifying that the use of pre-trained weights is crucial for large-scale models.

\begin{table}[t!]
\setlength{\tabcolsep}{5.7pt}
\caption{Comparison between trained (non-Bayesian) vanilla models with and without pre-trained weights on the Flowers dataset~\cite{flowers}. 
The weight decay hyperparameter is set to $10^{-4}$.
}
\vspace{-0.2em}
\centering
\begin{footnotesize}
\centering
\begin{tabular}{ccccc}
\toprule
\multirow{2}{*}{} & 
\multicolumn{2}{c}{ResNet-101} &
\multicolumn{2}{c}{ViT-L-32} 
\\
\cmidrule(lr){2-3} \cmidrule(lr){4-5} 
 & From-scratch & Pre-trained-warm-start & From-scratch & Pre-trained-warm-start
\\
\hline
Test error ($\%$)\Tstrut & 94.63 & 11.76 & 73.49 & 14.20 
\\
\bottomrule
\end{tabular}
\end{footnotesize}
\label{tab:nopretr_vs_pretr}
\end{table}

The overall test errors are shown in Table~\ref{tab:pets_errors}. For the detailed hyperparameters used in this experiment, please refer to our code. 
We see that the variational inference, MC-dropout, and SGLD models perform on par or better than deterministic models. 
The Laplace approximation performs reliably well with the posterior mean parameters (\texttt{nst}=0), but once we incorporate multiple posterior samples 
(\texttt{nst}=5) the test accuracy dropped significantly. We still investigate the precise reasons, but it might be due to the numerical issue in the diagonal empirical Fisher information estimate (e.g., one may as well put some larger regulariser in the denominator of (\ref{eq:la_posterior_emp_fisher}) for better numerical stability). Table~\ref{tab:pets_calibration} summarises the uncertainty quantification results.

\begin{table}[t!]
\setlength{\tabcolsep}{6.1pt}
\caption{Test errors ($\%$) of Bayesian adaptation/fene-tuning on the Pets dataset~\cite{pets}.
}
\vspace{-0.2em}
\centering
\begin{footnotesize}
\centering
\begin{tabular}{ccccccccccc}
\toprule
\multirow{2}{*}{} & 
\multicolumn{2}{c}{Vanilla} &
\multicolumn{2}{c}{VI} &
\multicolumn{2}{c}{MC-Dropout} &
\multicolumn{2}{c}{SGLD} &
\multicolumn{2}{c}{Laplace}
\\
\cmidrule(lr){2-3} \cmidrule(lr){4-5} \cmidrule(lr){6-7} \cmidrule(lr){8-9} \cmidrule(lr){10-11}
 & wd=0 & wd=$10^{-4}$ & nst=0 & nst=5 & nst=0 & nst=5 & nst=0 & nst=5 & nst=0 & nst=5
\\
\hline
ResNet-101\Tstrut & 10.03 & 10.03 & 10.03 & 9.27 & 10.03 & 9.65 & 9.21 & 9.24 & 10.19 & N/A
\\
ViT-L-32\Tstrut & 8.72 & 8.69 & 8.39 & 8.45 & 8.37 & 8.42 & 8.67 & 8.72 & 8.72 & N/A
\\
\bottomrule
\end{tabular}
\end{footnotesize}
\label{tab:pets_errors}
\end{table}

\begin{table}[t!]
\setlength{\tabcolsep}{5.8pt}
\caption{Uncertainty quantification of Bayesian foundation models on the Pets dataset~\cite{pets}.
}
\vspace{-0.2em}
\centering
\begin{footnotesize}
\centering
(a) ResNet-101 \\
\begin{tabular}{ccccccccccc}
\toprule
\multirow{2}{*}{} & 
\multicolumn{2}{c}{Vanilla} &
\multicolumn{2}{c}{VI} &
\multicolumn{2}{c}{MC-Dropout} &
\multicolumn{2}{c}{SGLD} &
\multicolumn{2}{c}{Laplace}
\\
\cmidrule(lr){2-3} \cmidrule(lr){4-5} \cmidrule(lr){6-7} \cmidrule(lr){8-9} \cmidrule(lr){10-11}
 & wd=0 & wd=$10^{-4}$ & nst=0 & nst=5 & nst=0 & nst=5 & nst=0 & nst=5 & nst=0 & nst=5
\\
\hline
ECE ($\%$)\Tstrut & 0.16 & 0.16 & 0.18 & 0.11 & 0.19 & 0.09 & 0.13 & 0.14 & 0.16 & N/A
\\
MCE ($\%$)\Tstrut & 10.02 & 12.24 & 11.95 & 8.53 & 14.76 & 8.10 & 12.17 & 15.29 & 15.71 & N/A
\\
NLL ($\times 10^{-2}$)\Tstrut & 33.73 & 33.59 & 33.53 & 31.04 & 33.38 & 31.68 & 31.70 & 31.40 & 33.79 & N/A
\\
\bottomrule
\end{tabular}
\\ \vspace{+0.5em}
(b) ViT-L-32 \\
\begin{tabular}{ccccccccccc}
\toprule
\multirow{2}{*}{} & 
\multicolumn{2}{c}{Vanilla} &
\multicolumn{2}{c}{VI} &
\multicolumn{2}{c}{MC-Dropout} &
\multicolumn{2}{c}{SGLD} &
\multicolumn{2}{c}{Laplace}
\\
\cmidrule(lr){2-3} \cmidrule(lr){4-5} \cmidrule(lr){6-7} \cmidrule(lr){8-9} \cmidrule(lr){10-11}
 & wd=0 & wd=$10^{-4}$ & nst=0 & nst=5 & nst=0 & nst=5 & nst=0 & nst=5 & nst=0 & nst=5
\\
\hline
ECE ($\%$)\Tstrut & 0.04 & 0.04 & 0.05 & 0.03 & 0.07 & 0.05 & 0.04 & 0.05 & 0.04 & N/A
\\
MCE ($\%$)\Tstrut & 11.55 & 12.43 & 9.59 & 7.19 & 13.38 & 8.98 & 11.09 & 10.78 & 11.55 & N/A
\\
NLL ($\times 10^{-2}$)\Tstrut & 26.72 & 26.69 & 25.65 & 25.99 & 25.64 & 25.79 & 27.01 & 26.99 & 26.72 & N/A
\\
\bottomrule
\end{tabular}
\end{footnotesize}
\label{tab:pets_calibration}
\end{table}

\begin{figure}
\begin{center}
%
\centering
\includegraphics[trim = 15mm 0mm 20mm 10mm, clip, scale=0.311]{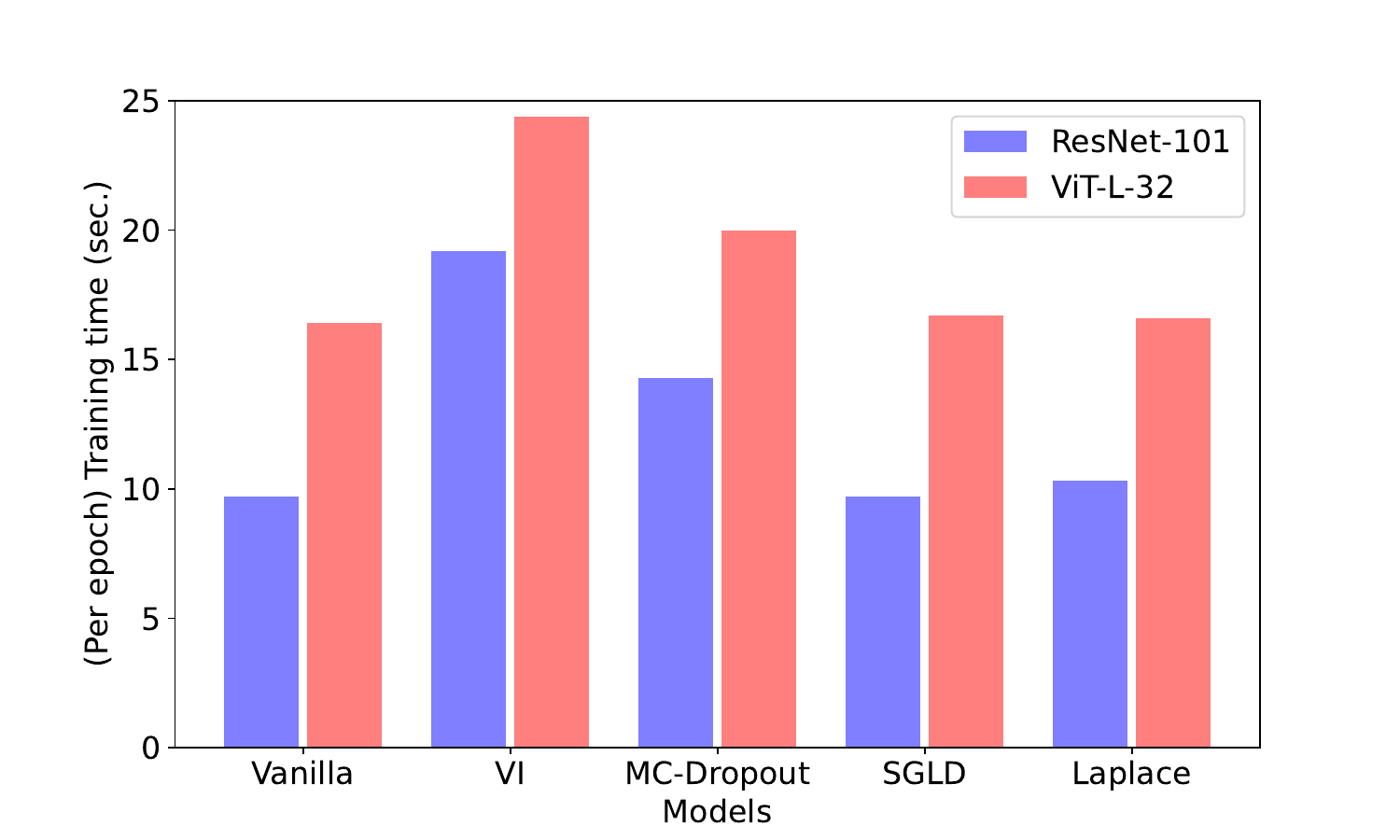} \ \ \ 
\includegraphics[trim = 15mm 0mm 20mm 10mm, clip, scale=0.311]{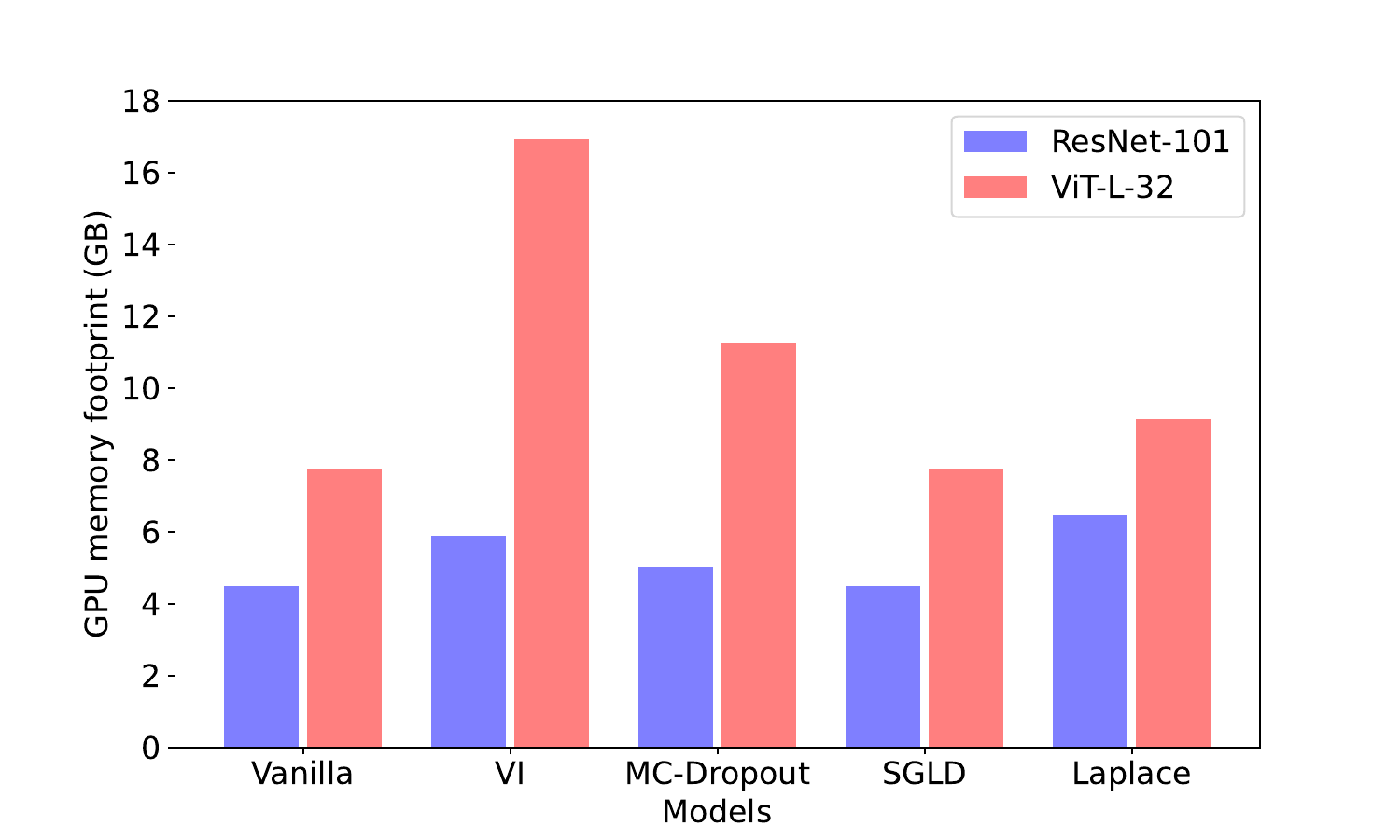}
\\
\ \ \ \ \ \ \ \ \ \ \ \ \ (a) Training time \ \ \ \ \ \ \ \ \ \ \ \ \ \ \ \ \ \ \ \ \ \ \ \ \ \ \ \ \ \ \ \ \ \ \ \ \ \ \ \ \ \ \ (b) GPU memory footprint
\end{center}
\vspace{-0.5em}
\caption{(a) (Per epoch) Training time (second) and (b) GPU memory footprint for different Bayesian models on the Flowers dataset~\cite{flowers}. 
}
\label{fig:complexity}
\end{figure}

\textbf{Computational overhead of Bayesian neural networks.}
One (often-believed) obstacle that prevents the Bayesian neural networks from being widely applied to large-scale foundation models in real world practice, is the computational overhead -- a sort of prejudice where one may well need to keep track of more parameters than deterministic models with increased training time. 
To clarify this, we actually compare the wall-clock training  times and memory footprints of the different Bayesian models against the deterministic vanilla SGD training in Fig.~\ref{fig:complexity}. 
We use a RTX-2080Ti machine for ResNet-101 and Tesla A100 for ViT-L-32, with single GPUs for both cases. 
As shown, all Bayesian approaches have tolerable overhead compared to base SGD models -- the over head is minor for SGLD and Laplace approximation; the worst-case overhead is at most two times of  base model's complexity. These results imply that the proposed Bayesian neural network library makes the Bayesianisation of large-scale foundation models viable.

\section{Conclusion}

We provide full implementation, without relying on other libraries, and easy-to-use demo codes for various Bayesian inference methods including: variational inference, MC-dropout, stochastic-gradient Langevin dynamics, and Laplace approximation. 
We also include the codes for evaluating Uncertainty Quantification measures provided (eg, ECE, MCE, Reliability plots, Negative log-likelihood), which can be used to report how well the uncertainty is captured in new models. Although we have tested the library with ResNet-101 and ViT-L-32, our library can be ready to be applicable to other Foundation Models such as LLAMA, RoBERTa, and Denoising Diffusion generative models without code modification at all. We also demonstrate that our code incurs minimal/acceptable use of extra computational resources (time and GPU memory). 




{
\small

{
\bibliographystyle{ieee_fullname}
\bibliography{main}   

\begin{thebibliography}{10}\itemsep=-1pt

\bibitem{bayes_by_backprop}
Charles Blundell, Julien Cornebise, Koray Kavukcuoglu, and Daan Wierstra.
\newblock {Weight Uncertainty in Neural Networks}.
\newblock In {\em International Conference on Machine Learning}, 2015.

\bibitem{sgmcmc1}
T. Chen, E.~B. Fox, and C. Guestrin.
\newblock {Stochastic gradient Hamiltonian Monte Carlo}.
\newblock {\em International Conference on Machine Learning}, 2014.

\bibitem{reliability1}
Morris~H. DeGroot and Stephen~E. Fienberg.
\newblock {The Comparison and Evaluation of Forecasters}.
\newblock {\em Journal of the Royal Statistical Society. Series D (The
  Statistician)}, 32(1/2):12--22, 1983.

\bibitem{deng2009imagenet}
Jia Deng, Wei Dong, Richard Socher, Li-Jia Li, Kai Li, and Li Fei-Fei.
\newblock {ImageNet: A large-scale hierarchical image database}.
\newblock In {\em IEEE Conference on Computer Vision and Pattern Recognition},
  2009.

\bibitem{sgmcmc2}
N. Ding, Y. Fang, R. Babbush, C. Chen, R.~D. Skeel, and H. Neven.
\newblock {Bayesian sampling using stochastic gradient thermostats}.
\newblock In {\em Advances in neural information processing systems}, 2014.

\bibitem{mc_dropout}
Yarin Gal and Zoubin Ghahramani.
\newblock {Dropout as a Bayesian Approximation: Representing Model Uncertainty
  in Deep Learning}.
\newblock In {\em International Conference on Machine Learning}, 2016.

\bibitem{Graves11}
A. Graves.
\newblock Practical variational inference for neural networks.
\newblock In {\em Advances in Neural Information Processing Systems}, 2011.

\bibitem{ece}
Chuan Guo, Geoff Pleiss, Yu Sun, and Kilian~Q. Weinberger.
\newblock {On Calibration of Modern Neural Networks}.
\newblock In {\em International Conference on Machine Learning}, 2017.

\bibitem{Martens14}
J. Martens.
\newblock {New insights and perspectives on the natural gradient method}.
\newblock {\em arXiv preprint arXiv:1412.1193}, 2014.

\bibitem{reliability2}
Alexandru Niculescu-Mizil and Rich Caruana.
\newblock {Predicting good probabilities with supervised learning}.
\newblock In {\em International Conference on Machine Learning}, 2005.

\bibitem{flowers}
M-E. Nilsback and A. Zisserman.
\newblock {Automated flower classification over a large number of classes}.
\newblock In {\em Proceedings of the Indian Conference on Computer Vision,
  Graphics and Image Processing}, 2008.

\bibitem{pets}
O.~M. Parkhi, A. Vedaldi, A. Zisserman, and C.~V. Jawahar.
\newblock {Cats and Dogs}.
\newblock In {\em IEEE Conference on Computer Vision and Pattern Recognition},
  2012.

\bibitem{kfac_la}
Hippolyt Ritter, Aleksandar Botev, and David Barber.
\newblock {A Scalable Laplace Approximation for Neural Networks}.
\newblock In {\em International Conference on Learning Representations}, 2018.

\bibitem{Schraudolph02}
N.~N. Schraudolph.
\newblock Fast curvature matrix-vector products for second-order gradient
  descent.
\newblock {\em Neural computation}, 14(7):1723--1738, 2002.

\bibitem{sgld}
Max Welling and Yee~Whye Teh.
\newblock {Bayesian Learning via Stochastic Gradient Langevin Dynamics}.
\newblock In {\em International Conference on Machine Learning}, 2011.

\end{thebibliography}
}



}


\end{document}